\documentclass{article}

\usepackage[colorlinks]{hyperref}       % hyperlinks
\usepackage{amsfonts}       % blackboard math symbols
\usepackage{graphicx}
\usepackage{multirow}
\usepackage{multicol}
\usepackage{subcaption}
\usepackage{arxiv}
\usepackage{euscript}
\usepackage{amsmath} 

\newcommand{\ModelName}{ColonFormer}    
\title{\ModelName: An Efficient Transformer based Method for Colon Polyp Segmentation}

\author{{Nguyen Thanh Duc} \\
School of Information and Communication Technology\\
Hanoi University of Science and Technology\\
\texttt{duc.nt170058@sis.hust.edu.vn} \\
\And
{Nguyen Thi Oanh} \\
School of Information and Communication Technology\\
Hanoi University of Science and Technology\\
\texttt{oanhnt@soict.hust.edu.vn} \\
\And
{Nguyen Thi Thuy} \\
Faculty of Infomation Technology\\
Vietnam National University of Agriculture\\
\texttt{ntthuy@vnua.edu.vn} \\
\And
{Tran Minh Triet} \\
University of Science\\
Vietnam National University Ho Chi Minh City\\
\texttt{tmtriet@fit.hcmus.edu.vn} \\
\And
{Dinh Viet Sang} \thanks{Corresponding author}\\
School of Information and Communication Technology\\
Hanoi University of Science and Technology\\
\texttt{sangdv@soict.hust.edu.vn} \\
}

\renewcommand{\shorttitle}{\ModelName: An Efficient Transformer based Method for Colon Polyp Segmentation}

\begin{document}
\maketitle

\begin{abstract}
Identifying polyps is challenging for automatic analysis of endoscopic images in computer-aided clinical support systems.
Models based on convolutional networks (CNN), transformers, and their combinations have been proposed to segment polyps with promising results. However, those approaches have limitations either in modeling the local appearance of the polyps only or lack of multi-level feature representation for spatial dependency in the decoding process.  
This paper proposes a novel network, namely ColonFormer, to address these limitations.
ColonFormer is an encoder-decoder architecture capable of modeling long-range semantic information at both encoder and decoder branches. The encoder is a lightweight architecture based on transformers for modeling global semantic relations at multi scales. The decoder is a hierarchical network structure designed for learning multi-level features to enrich feature representation. Besides, a refinement module is added with a new skip connection technique to refine the boundary of polyp objects in the global map for accurate segmentation.
Extensive experiments have been conducted on five popular benchmark datasets for polyp segmentation, including Kvasir, CVC-Clinic DB, CVC-ColonDB, CVC-T, and ETIS-Larib. Experimental results show that our ColonFormer outperforms other state-of-the-art methods on all benchmark datasets.
\end{abstract}

% \begin{keywords}
% Semantic segmentation, deep learning, encoder-decoder network, polyp segmentation, colonoscopy
% \end{keywords}

% keywords can be removed

\section{Introduction}
\label{sec:introduction}
% NEED PARAPHRASE
Colorectal cancer (CRC) is among the most common types of cancer worldwide, causing over 694,000 fatalities each year \cite{bernal2017comparative}.
The most common cause of CRC is colon polyps, particularly adenomas with high-grade dysplasia. %\cite{gschwantler2002high}.
According to a longitudinal study \cite{corley2014adenoma}, every 1$\%$ increase in adenoma detection rate is linked to a 3$\%$ reduction in the risk of colon cancer. As a result, detecting and removing polyps at an early stage is critical for cancer prevention and treatment. Therefore, colonoscopy is regarded as the standard tool for detecting colon adenomas and colorectal cancer.
%colon screening and is recommended procedure in many different societies' guidelines. % \cite{issa2017colorectal}. 
In practice, overburdened healthcare systems, particularly in low-resource settings, may result in shorter endoscopy durations and more missed polyps.  %\cite{lee2008adequate, armin2015visibility}. 
According to a literature review, the proportion of colon polyps missing during endoscopies could range from 20 to 47 percent \cite{leufkens2012factors}. This may lead to high associated risk factors in patients. Therefore, research in developing computer-aided tools to assist endoscopists in endoscopy procedures is an essential need. % , both in terms of training and implementation in clinical practice.

Advancements in artificial intelligence and deep learning have changed the landscape of such systems. Attempts have been made to develop learning algorithms to deploy in computer-aided diagnostic (CAD) systems for the automatic detection and prediction of polyps, which could benefit clinicians in detecting lesions and lower the miss detection rate \cite{mesejo2016computer,zhou2019951e,kudo2019artificial}. 
Deep neural networks have shown great potential in assisting colon polyp detection in several retrospective investigations and diagnoses. 
%Endoscopists should be supported in lesion detection, diagnosis, and quality assurance by the CAD system. Better computer support systems will help improve lesion detection rates, optimize strategies during endoscopy for high-risk lesions, and increase clinics' capacity while preserving diagnostic quality \cite{chen2018accurate, bisschops2019advanced}.
A CAD system can support endoscopists in improving lesion detection rates, optimizing strategies during endoscopy for high-risk lesions, and increasing clinics' capacity while preserving diagnostic quality \cite{chen2018accurate, bisschops2019advanced}.

Despite progress in machine learning and computer vision research, automatic polyp segmentation remains a challenging problem. Polyps are caused by abnormal cell growth in the human colon, meaning their appearances have strong relationships with the surroundings. Images of polyps come in various shapes, sizes, textures, and colors. In addition, the boundary between polyps and their surrounding mucosa is not always apparent during colonoscopy, especially in different lighting modes and in cases of flat lesions or unclean bowel preparation. These cause a lot of uncertainty for the learning models for polyp segmentation.

In recent years, the most widely used methods for image segmentation in general and polyp segmentation, in particular, are based on Convolutional Neural Networks (CNNs). Most segmentation models use a UNet-based architecture containing an encoder and a decoder, which are often built up from convolutional layers. Despite being widely used for segmentation tasks with impressive performance, CNNs pose certain limitations: They can only capture local information while ignoring spatial context and global information due to the limited receptive field. Furthermore, it was shown that CNNs act like a series of high-pass filters and favor high-frequency information.
% TODO: Bổ sung trích dẫn cho nhận định trên
% Integrating the self-attention mechanism into CNNs is one solution to enhance performance. Combining CNN with self-attention processes improves the ability to model global interactions. This method yields positive outcomes.

Transformer \cite{transformer2017} is a recently proposed deep neural network architecture that models the global interactions among input components using attention mechanisms. While initially designed to tackle natural language and speech processing problems, Transformers have significantly impacted computer vision in recent years. In contrast to CNNs, self-attention layers in Transformers work as low-pass filters, and they can effectively capture long-range dependency. Therefore, combining the strengths of convolutional and self-attention layers can increase the representation power of deep networks. 
%ViT can only work with fixed resolutions to generate single low-resolution feature maps, making it limited in producing informative features at different scales and resolutions.
% TODO
Very recently, there has been fast-growing interest in using Transformers for semantic image segmentation \cite{swin,wang2021pyramid, ranftl2021vision, zheng2021rethinking}. These methods use well-known encoder-decoder architectures wherein Transformers and CNNs are combined in various settings. The works in \cite{swin,wang2021pyramid, ranftl2021vision} proposed Transformer-CNN architectures, in which a Transformer is used as an encoder, and a traditional CNN is used as a decoder. The hybrid architecture of Transformers and CNN has been proposed in  \cite{zheng2021rethinking}, in which the decoder is a traditional CNN or a Transformer, while the encoder is a combination of CNN and Transformer layers. 

Inspired by these approaches for modeling multi-scale and multi-level features, we propose a new Transformers-based network called \ModelName. The main design of our \ModelName also contains a transformer encoder and a CNN decoder, but our approach is different from the models mentioned above in several ways. In  \ModelName, the encoder is a hierarchically structured lightweight Transformer for learning multi-scale features. The decoder is a hierarchical pyramid structure with the capability of learning from heterogeneous data containing feature maps extracted from encoder blocks at different scales and subregions. Besides, a refinement module is proposed for further improving the segmentation accuracy on hard regions and small polyps. 

Our main contributions include: 
\begin{itemize}
\item A novel deep neural network, namely \ModelName, that integrates a hierarchical Transformer and a hierarchical pyramid CNN in a unified architecture for efficient and accurate polyp segmentation;
\item An improved refinement technique using a newly proposed residual axial attention module for feature fusion and smoothing aiming at improving the segmentation accuracy;
\item A set of experiments on five standard benchmark datasets for polyp segmentation (Kvasir, CVC-Clinic DB, CVC-ColonDB, CVC-T, and ETIS-Larib) and comparisons of the effectiveness of \ModelName to current state-of-the-art methods.
\end{itemize}

\begin{figure*}[ht!]
\centering
\includegraphics[width=0.8\textwidth]{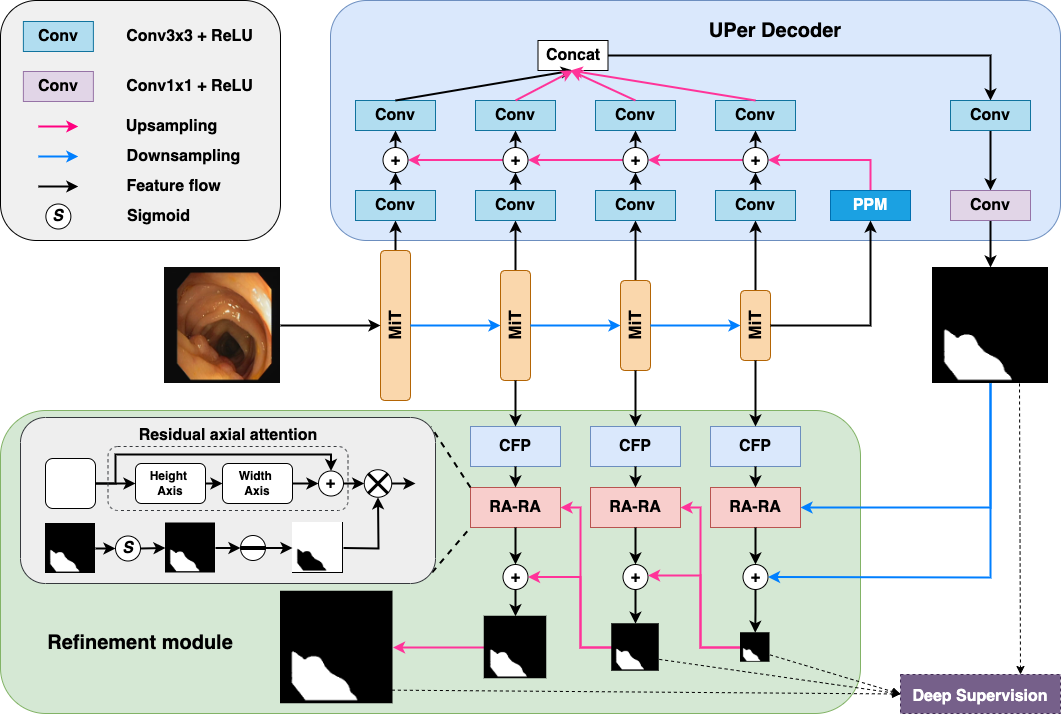} 
\caption{The overall architecture of our \ModelName contains three components: an encoder, a decoder, and a refinement module. The encoder is based on the Mix Transformer backbone. The decoder starts with a pyramid pooling module (PPM), where its outputs are combined layer-wise with the output feature maps of the encoder at multi levels to produce a global map. The refinement module aims to gradually refine the boundary of the global map to yield the final accurate segmentation. 
Besides this predicted output, the global map and two intermediate maps are also passed into the training loss in a deep supervision manner. Before calculating the training loss, all refined maps are upsampled back to the original image input size.}
\label{fig:colonformer}
\end{figure*}

The rest of the paper is organized as follows. We provide a brief review of related works in Section \ref{sec:related}. The \ModelName architecture is described in Section \ref{sec:propose}. Section \ref{sec:experiment} presents our experiments and results. Finally, we conclude the paper and highlight future works in Section \ref{sec:conclude}.

\section{Related Work}
\label{sec:related}
In this section, we briefly review common methods and techniques that have been developed for polyp segmentation. We review CNN architectures and their variants, especially UNet models, in medical image segmentation. Then we investigate the attention mechanism as a promising technique that boosts the capability of a deep neural network in learning feature representation. Finally, vision Transformer and its applications in polyp segmentation and medical image processing are investigated.

\subsection{Convolutional Neural Networks}
CNNs are one of the most widely used deep neural network architectures, especially in computer vision. A deep CNN extracts features on multiple layers with increasing levels of abstraction. Low-level features with high resolutions represent spatial details, while high-level features with low resolutions represent rich semantic information. CNNs are especially powerful in image analysis as they can extract highly informative and valuable features. 

UNet \cite{unet} is a pioneering CNN architecture for medical image segmentation. UNet consists of an encoder and a decoder. The encoder includes convolutional, pooling layers for feature extraction, and the decoder uses upsampling (or deconvolutional) and convolutional layers for yielding the final segmentation prediction. Later works attempted to improve UNet by introducing skip connections, which alleviate information loss caused by stacking multiple convolutional layers. However, retaining information from low-level may introduce noisy signals that degrade the performance. UNet variants such as UNet++ \cite{unet++} and DoubleUnet \cite{doubleunet} have achieved stellar results on segmentation benchmarks. UNet++ is constructed as an ensemble of nested UNets of varying depths, which partially share an encoder and jointly learn using deep supervision. DoubleUNet stacks two UNet blocks and uses ASPP \cite{aspp}, and SE blocks \cite{seblock} to enhance the feature representation.

UNet encoders often use an existing pretrained architecture, also known as the backbone. Widely used backbones include VGG \cite{vgg}, MobileNet \cite{mobilenet}, ResNet \cite{resnet}, DenseNet \cite{densenet}, etc. PraNet \cite{pranet} uses Res2Net as the backbone, while AG-CuResNeSt \cite{agcuresnest} uses ResNeSt. Meanwhile HarDNet-MSEG \cite{hardnet_mseg}, NeoUNet \cite{ngoc2021neounet} and BlazeNeo \cite{an2022blazeneo} use HarDNet, an improvement of DenseNet to extract features.

\subsection{Attention Mechanism}
The attention mechanism is a widely used technique to help deep neural networks learn better feature representations, especially on highly variant inputs. 
Oktay et al. \cite{unetag} proposed an Attention Gate module for UNet, which helps the model focus on necessary information while preserving computational efficiency. AG-ResUNet++ \cite{hung2021ag} integrates the attention gates with the ResNet backbone to improve UNet++ \cite{unet++} for polyp segmentation. PraNet \cite{pranet} uses the Reverse Attention module \cite{ra}, which enforces focus on the boundary between a polyp and its surroundings. In general, most CNNs and neural networks can benefit by adding attention modules. However, even with these attention mechanisms, CNNs are limited by the locality of convolution operations. This limitation makes them difficult to model natural long-range spatial dependencies between input segments.

\begin{figure*}[ht!]
\centering
\includegraphics[width=\textwidth]{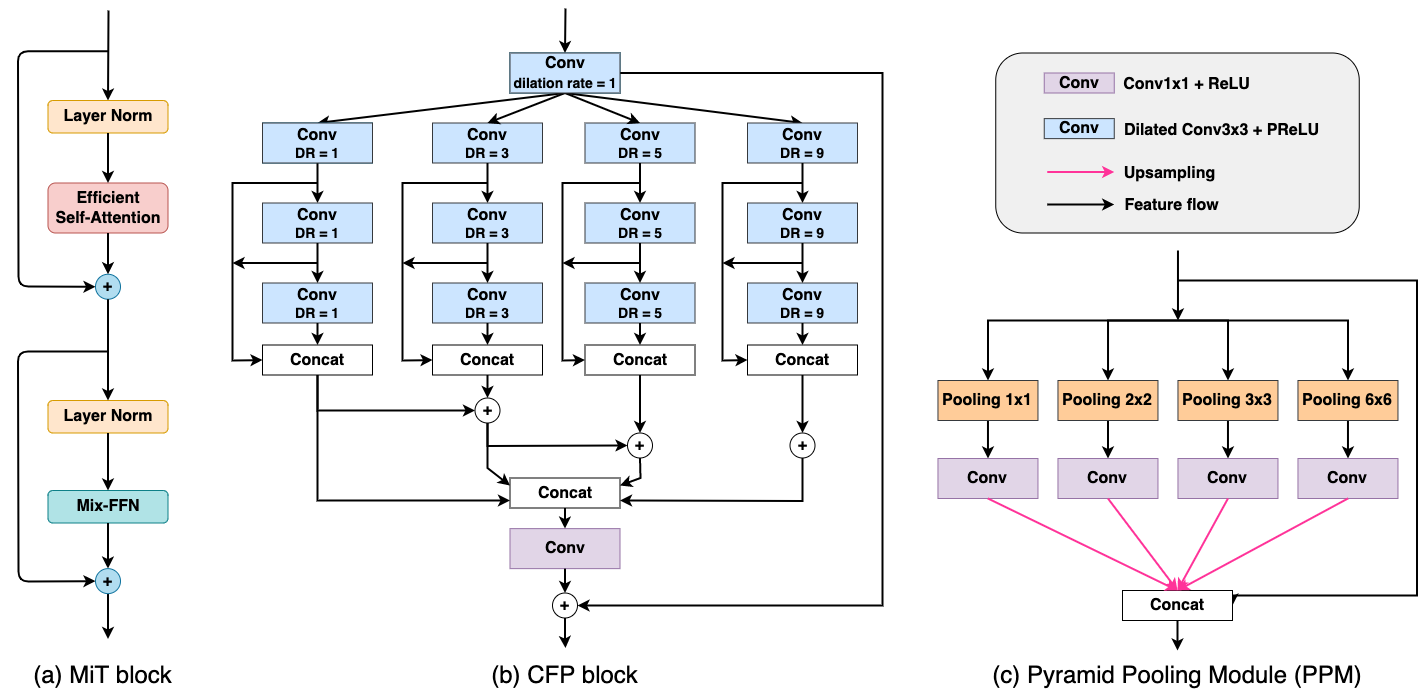} 
\caption{The architecture of three neural blocks used in our \ModelName. The left block (a) is the Mix Transformer block  \cite{segformer}. The middle block (b) is the channel-wise feature pyramid block. The pyramid pooling module is shown on the right (c). Here DR stands for dilation rate.}
\label{fig:cfp-mit-ppm}
\end{figure*}

\subsection{Vision Transformer}
Transformer \cite{transformer2017} is a highly influential deep neural network architecture, originally proposed to solve natural language processing and similar problems. While the original Transformer architecture is not very well suited for image analysis, attempts have been proposed to leverage its advantages for computer vision through some modifications. Vision Transformer (ViT) \cite{vit} was the first successful application of Transformers for computer vision. ViT splits an image into patches and processes them as sequential tokens. This method greatly reduces computational costs and allows Transformers to work with large images feasibly. 

A major issue of ViT is that it requires extensive datasets for training to remain effective while being severely limited when trained on small datasets. Such property hinders its usage in problems such as medical image analysis, including polyp segmentation, where data is typically scarce. The Kvasir dataset, for example, contains just 1000 images and their corresponding ground truth, despite being the largest public image dataset of the gastrointestinal tract for polyp segmentation. 

Recent works have attempted to further enhance ViT in several ways. DeiT \cite{deit} introduces a data-efficient training strategy combined with a distillation approach, which helps improve the performance when training on small datasets. Swin Transformer \cite{swin} redesigned the encoder for Transformers. The Swin Transformer encoder computes self-attention among a collection of adjacent patches within a sliding window. Patches are merged every few blocks, reducing the number of tokens and forming a multi-resolution token hierarchy similar to convolutional blocks. SegFormer \cite{segformer} is another hierarchical Transformer design, where patches are merged with overlap and preserving local continuity around patches. The authors also introduced Efficient Self-Attention, a modified attention mechanism for reducing computational complexity, and Mix-FFN for better positional information.

Both TransUNet \cite{transunet} and TransFuse \cite{transfuse} models have been developed based on Transformers for polyp segmentation and yielded promising results. TransUNet uses a Transformer-based network with a hybrid ViT encoder and upsampling CNN decoder. The Hybrid ViT stacks the CNN and Transformer together, leading to high computational costs. TransFuse addressed this problem by using a parallel architecture. Both models use the attention gate mechanism \cite{unetag} and a so-called BiFusion Module. %, which include SE Blocks \cite{seblock} and CBAM Blocks \cite{cbam}. 
These components make the network architecture large and highly complex.

While there have been promising results in using Transformers to develop networks for polyp segmentation, there is plenty of room for improvement in this direction. Most notably, reduced network size and latency can greatly benefit downstream applications. In addition, improved accuracy and robustness can also be achieved with more optimized architectures. This paper seeks to design a Transformer-based architecture that achieves these goals. 

\section{\ModelName}
\label{sec:propose}
Fig.~\ref{fig:colonformer} depicts the overall architecture of our proposed network, \ModelName. The network consists of a hybrid encoder, a decoder, and a refinement module. We will describe each component in detail in the following sections.

\subsection{Encoder}

A hierarchically structured model that can extract coarse-to-fine features at multi-scale and multi-level is desired for semantic segmentation. Our model uses Mix Transformer (MiT) proposed in \cite{segformer} as the encoder backbone. MiT is a hierarchical Transformer encoder that can represent both high-resolution coarse and low-resolution fine features. Assume $X \in  \mathbb{R}^{H \times W \times C}$ denotes the input image. MiT generates the CNN-like multi-level features $F_i$. The hierarchical feature map $F_i$ has the resolution of ${\frac{H}{2^{i+1}} \times \frac{W}{2^{i+1}} \times C_i}$, where $i \in \{1,2,3,4\}$ and $C_i$ is ascending. The hierarchical feature representation is brought by the overlapped patch merging. After several Transformer blocks (Fig.~\ref{fig:cfp-mit-ppm}a), a kernel with a stride smaller than kernel size is used to divide the feature map into overlapping patches. Such an overlapping patch merging process ensures the local continuity around those patches.  %By striding the overlapped kernel, it can preserve the local continuity around those patches. 

Like other Transformer blocks, MiT blocks contain three main parts: Multi-head Self-Attention (MHSA) layers, Feed Forward Network (FFN), and Layer Norm. The MHSA is improved into Efficient Self-Attention, where the number of keys is decreased by a factor of $R$ to reduce the computational complexity of self-attention layers. Another reason that we decide to choose MiT is the Mix-FFN. Instead of using the positional encoding (PE) as ViT, a $3\times3$ convolution kernel is integrated into FFN. Since the resolution of PE is fixed, it can not utilize the positional information of the pretrained dataset like ImageNet when test resolution differs from the training one. In such cases, ViT \cite{vit} suggests interpolating the PE, which can lead to a drop in accuracy. In contrast, arguing that convolutional layers are adequate for providing location information for Transformer, MiT directly uses a 3x3 convolutional layer for positional encoding. MiT has a series of variants, from MiT-B1 to MiT-B5, with the same architecture but different sizes. We name the variants of our model as {\textbf{{\ModelName-XS}}, {\textbf{\ModelName-S}}, {\textbf{\ModelName-L}}, {\textbf{\ModelName-XL}}, {\textbf{\ModelName-XXL}}, corresponding to different MiT backbone scales from MiT-B1 to MiT-B5, respectively.
According to ablation study described later in Section~\ref{sec:ablation_studies}, we found that {\ModelName-S} and {\ModelName-L} achieve the best results. Therefore, we mostly use {\ModelName-S} and {\ModelName-L} for comparison with other state-of-the-arts in all experiments except where it is specified otherwise. 

% \begin{figure}[ht!]
%     \centering
%     \includegraphics[scale=0.45]{Images/segformer_block.png} 
%     \caption{Mix Transformer block.}
%     \label{fig:segformer_block}
% \end{figure}

\subsection{Decoder}
% In the previous CNN models like UNet \cite{unet}, UNet++ \cite{unet++}, a sequential convolution decoder that consists of upsampling layers and convolution layers is used. PraNet \cite{pranet} and HarDNet-MSEG \cite{hardnet_mseg} use Parallel Partial Decoder (PPD) \cite{ppd}. PPD contains parallel convolution branches. It selects and aggregates high-level features. Each branch uses different kernel sizes, so the model can see feature maps in different views, which helps the selective information become richer and more diverse. Original SegFormer \cite{segformer} uses a lightweight MLP Decoder, which only contains MLP layers. All feature maps extracted from the encoder go through MLP layers to unify the channel dimension. After that, they are resized to the same scale and concatenated together. Finally, MLP layers are used to fuse the concatenated features and predict the final map. Although using MLP for the decoder part is quite fast and accurate, while we experiment, we found that there is a more efficient method. \\
%\textcolor{red}
%{Lan: nen mo ta ki hon cho UPer Decoder}

In order to further capture global context information, the feature maps extracted from the final block of the encoder are first processed by a Pyramid Pooling Module (PPM) \cite{ppm} before being passed through the decoder blocks. The PPM simultaneously produces multi-scale outputs of the input feature map via a pyramid of pooling layers. The resulting feature maps, which form a hierarchy of features containing information at different scales and sub-regions, are then concatenated to produce an efficient prior global representation. Fig.~\ref{fig:cfp-mit-ppm}c depicts the Pyramid Pooling Module in detail.

\ModelName uses a decoder architecture inspired by UPerNet \cite{upernet}, which we denote as UPer Decoder. The decoder gradually fuses the prior global map produced by the PPM with multi-scale feature maps yielded by the MiT backbone. We suppose that applying convolutional layers to the feature maps of the MiT backbone is necessary since such layers can condense the information by emphasizing the coherence between neighboring elements and thus enhancing the resulting semantic map. 

\subsection{Refinement Module}
\label{sec:refinement_module}
The decoder's outputs are further processed by a refinement module to achieve more precise and complete prediction maps. The refinement module consists of Channel-wise Feature Pyramid (CFP) module \cite{cfpnet} (Fig.~\ref{fig:cfp-mit-ppm}b) and our novel  Reverse Attention module enhanced by a new residual axial attention block for incremental correction of polyp boundary \cite{ra, aa}.

%Doctors in a clinical context initially find the polyp location roughly before carefully inspecting local tissues to precisely label the polyp. 
In the parallel reverse attention network architecture \cite{pranet}, the global map is derived from the deepest CNN layer, so it does not have many structural details and hence can present only rough locations of the polyp tissues. The proposed strategy to recover precise location and label is to exploit complementary regions and details in a sequential manner by removing previously estimated polyp regions from high-level side-output features, where the current estimation is up-sampled from the deeper layer. By using Reverse Attention, a coarse saliency map is guided to sequentially discover complement object regions and details by erasing the current predicted salient regions from side-output features. The current prediction is upsampled from its deeper layer. This erasing approach can refine the imprecise and coarse estimation into an accurate and complete prediction map.
It was shown that self-attention layers in the MiT backbone work like low-pass filters. Therefore, we argue that using convolutional layers is important for the refinement module since such layers favor high-frequency components and can provide richer edge information for the boundary correction.

Inspired by CaraNet \cite{caranet}, we use Channel-wise Feature Pyramid (CFP) to extract features from the encoder in multi-scale views. As depicted in Fig.~\ref{fig:cfp-mit-ppm}b, the CFP module has $K=4$ branches with different dilation rates that allow them to capture information at multiple scales. However, a direct concatenation of all branches could lead to some unwanted checkerboard or gridding artifacts that significantly impact the quality of the following boundary correction. In order to avoid this issue, the CFP module combines these branches step by step to build a final accurate feature map to correct the polyp boundaries.  

CaraNet \cite{caranet} also enhanced the Reverse Attention module by an axial attention block, which is a straightforward generalization of self-attention that naturally aligns with the multiple dimensions of the tensors. This module is supposed to filter the necessary information for the refinement process. However, axial attention may not always be good for the network since it can accidentally eliminate important edge information. Therefore, we propose to relax this mechanism using an additional residual connection, which allows the network to omit the axial attention layers when required and thus facilitates the learning process. The novel refinement module is called Residual Axial Reverse Attention (RA-RA). We experimentally found that utilizing the RA-RA module up to the finest feature map does not help refine the polyp boundary better. Hence, we propose to use just three RA-RA blocks, as shown in Fig.~\ref{fig:colonformer}. The effectiveness of the RA-RA module is investigated in detail in Section~\ref{sec:ablation_studies}.

% \begin{figure}[ht!]
%     \centering
%     \includegraphics[scale=0.45]{Images/ara.png} 
%     \caption{Axial Reverse Attention Module.}
%     \label{fig:ara}
% \end{figure}

\subsection{Loss Function}
\ModelName uses a compound loss combining the weighted focal loss and weighted IoU loss to train the model. The weighted focal loss is a distribution-based loss that treats every pixel individually. In contrast, the weighted IoU loss is a region-based loss that considers the relationships between neighboring pixels.

Image pixels can be easy to be correctly recognized. However, some pixels, such as those on the edge regions, may be harder to learn. Thus, the model should pay more attention to more challenging samples. In other words, some image pixels may be more important than others in contributing to the learning process. We represent the importance of pixel $(i, j)$ by a weight $\beta_{ij}$. As suggested in \cite{f3net}, the weight $\beta_{ij}$ for pixel $(i, j)$ is defined as the difference between the center pixel and its neighbors: 
\begin{equation}
\beta_{ij} = \Big | \frac{\sum_{m,n \in \EuScript{N}_{ij}}^{}g_{mn}}{|\EuScript{N}_{ij}|} - g_{ij} \Big |
\end{equation}
where $\EuScript{N}_{ij}$ represents the area of $31 \times 31$ size surrounding the pixel $(i,j)$, and $g_{ij} \in \{0,1\}$ is the true label of pixel $(i, j)$. A large value of $\beta_{ij}$ indicates a pixel with considerable distinction from its vicinity, i.e., pixels at polyp edges. Such a weighting scheme enforces the model to focus more on the boundary regions. 

Assume that $p_{ij}$ is the prediction probability of the pixel $(i,j)$ belonging the polyp class. Let us define $q_{ij}$ as:
\begin{equation}
q_{ij} = \left\{\begin{matrix}
p_{ij}, & \mbox{if $g_{ij}$ = 1} & \\ 
1-p_{ij}, & \mbox{otherwise} &
\end{matrix}\right.
\end{equation}

As polyp segmentation is a problem with highly imbalanced data, focal loss is employed to deal with class imbalance during training. It integrates a modulating term in order to focus learning on hard pixels. The weighted focal loss is then defined as follows:
\begin{equation}
\label{eq:wfocal}
\mathcal{L}_{wfocal} = -\frac{\sum_{i=1}^{H}\sum_{j=1}^{W}(1+\lambda\beta{ij})\alpha(1-q_{ij})^\gamma\log(q_{ij})}{\sum_{i=1}^{H}\sum_{j=1}^{W}(1 +\lambda\beta_{ij})}
\end{equation}
where $\alpha, \gamma$ are tunable hyperparameters.

The weighted IoU loss is defined as follows:
\begin{equation}
\label{eq:wiou}
\mathcal{L}_{wiou} = 1 - \frac{\sum_{i=1}^{H}\sum_{j=1}^{W}(g_{ij}*p_{ij})*(1+\lambda\beta_{ij})}{\sum_{i=1}^{H}\sum_{j=1}^{W}(g_{ij} + p_{ij} - g_{ij}*p_{ij})*(1+\lambda\beta_{ij})}
\end{equation}
where $\lambda$ is a hyperparameter to adjust the impact of importance weights $\beta_{ij}$.

The total loss of our \ModelName is calculated as: 
\begin{equation}
\mathcal{L}_{total} = \frac{\mathcal{L}_{wfocal} + \mathcal{L}_{wiou}}{2}
\label{eq:totalloss}
\end{equation}

The total loss in Eq.~(\ref{eq:totalloss}) is applied to train our model for multi-scale outputs as shown in Fig.~\ref{fig:colonformer}. The final loss is the sum of all total losses computed at different output levels. Note that each output is upsampled back to the original size of the image's ground truth before the losses are evaluated.

% \textcolor{red}{thieu giai thich $p_{ij}$ CT can check lai}

\section{Experiments}
\label{sec:experiment}

\begin{table*}
\caption{Performance comparison of different methods on the Kvasir, ClinicDB, ColonDB, CVC-T and ETIS-Larib test sets. All results of \ModelName are averaged over five runs.}
\centering
{\renewcommand{\arraystretch}{1.2}
\begin{tabular}{c|cc|cc|cc|cc|cc}
\hline
Method & \multicolumn{2}{c|}{Kvasir} & \multicolumn{2}{c|}{CVC-ClinicDB} & \multicolumn{2}{c|}{CVC-ColonDB} & \multicolumn{2}{c|}{CVC-T} & \multicolumn{2}{c}{ETIS-Larib}  \\
\cline{2-11}
& mDice & mIOU               & mDice & mIOU                 & mDice & mIOU                & mDice & mIOU                   & mDice & mIOU              \\
\hline
\hline
UNet \cite{unet}           & 0.818 & 0.746 & 0.823 & 0.750 & 0.512 & 0.444 & 0.710 & 0.627  & 0.398 & 0.335 \\
UNet++ \cite{unet++}         & 0.821 & 0.743 & 0.794 & 0.729 & 0.483 & 0.410 & 0.707 & 0.624  & 0.401 & 0.344 \\
SFA   \cite{sfa}                      & 0.723 & 0.611 & 0.700 & 0.607 & 0.469 & 0.347 & 0.297 & 0.217 & 0.467 & 0.329\\
PraNet \cite{pranet}         & 0.898 & 0.840 & 0.899 & 0.849 & 0.709 & 0.640 & 0.871 & 0.797  & 0.628 & 0.567 \\
HarDNet-MSEG \cite{hardnet_mseg}  & 0.912 & 0.857 & 0.932 & 0.882 & 0.731 & 0.660 & 0.887 & 0.821  & 0.677 & 0.613 \\
CaraNet \cite{caranet}          & 0.918 & 0.865 & 0.936 & 0.887 & 0.773 & 0.689 & 0.903 & 0.838 & 0.747 & 0.672\\
TransUNet \cite{transunet}   & 0.913 & 0.857 & 0.935 & 0.887 & 0.781 & 0.699 & 0.893 & 0.824 & 0.731 & 0.660   \\
%TransFuse-S \cite{transfuse}    & 0.918 & 0.868 & 0.918 & 0.868 & 0.773 & 0.696 & 0.902 & 0.833  & 0.733 & 0.659 \\
%TransFuse-L \cite{transfuse}    & 0.918 & 0.868 & 0.934 & 0.886 & 0.744 & 0.676 & 0.904 & 0.838  & 0.737 & 0.661 \\
TransFuse-L* \cite{transfuse}   & 0.920 & 0.870 & \textbf{0.942} & \textbf{0.897} & \underline{\textit{0.781}} & 0.706 & \underline{\textit{0.894}} & \underline{\textit{0.826}} & 0.737 & 0.663\\
%\hline
\textit{\textbf{\ModelName-S~(Ours)}} & \textbf{0.927} & \textbf{0.877} & \underline{\textit{0.932}} & 0.883 & \textbf{0.811} & \underline{\textit{0.730}} & \underline{\textit{0.894}} & \underline{\textit{0.826}}  & \underline{\textit{0.789}} & \underline{\textit{0.711}} \\
\textit{\textbf{\ModelName-L~(Ours)}} & \underline{\textit{0.924}} & \underline{\textit{0.876}} & \underline{\textit{0.932}} & \underline{\textit{0.884}} & \textbf{0.811} & \textbf{0.733} & \textbf{0.906} & \textbf{0.842} & \textbf{0.801} & \textbf{0.722} \\[2pt]
\hline
\end{tabular}
}
\label{tab:sota}
\end{table*}

\begin{table*}[ht!]
\caption{Performance comparison of different methods on 5-fold cross-validation of the CVC-ClinicDB and Kvasir datasets. All results are averaged over 5 folds.}
\centering
{\renewcommand{\arraystretch}{1.2}
\begin{tabular}{c|c|cccc}
\hline
Dataset & Method & mDice & mIOU  & Recall & Precision  \\
\hline
\hline
\multirow{9}{*}{\rotatebox[origin=c]{90}{ClinicDB}} 
& UNet \cite{unet}                   & -          & 0.792              & -          & -           \\
& MultiResUNet \cite{multiresunet}& -          & 0.849              & -          & -           \\
& ResUNet++ \cite{resunet++}  & $0.815 \pm 0.018$ & $0.736 \pm 0.017$ & $0.832\pm 0.018$ &  $0.830 \pm 0.020$ \\
& DoubleUNet \cite{doubleunet} & $0.920 \pm 0.018$ & $0.866 \pm 0.025$ & $0.922 \pm 0.027$ & $0.928 \pm 0.017$ \\
& DDANet \cite{ddanet} & $0.860 \pm 0.014$ & $0.786 \pm 0.017$ & $0.858 \pm 0.023$ & $0.892 \pm 0.014$ \\
& ColonSegNet \cite{colonsegnet} & $0.817 \pm 0.020$ & $0.873 \pm 0.024$ & $0.926 \pm 0.025$ & $0.933 \pm 0.014$ \\
& HarDNet-MSEG \cite{hardnet_mseg} & $0.923 \pm 0.020$ & $0.873 \pm 0.024$ & $0.926 \pm 0.025$ & $0.933 \pm 0.014$ \\
& PraNet \cite{pranet} & $0.933 \pm 0.012$ & $0.884 \pm 0.015$ & $0.940 \pm 0.005$ & $0.937 \pm 0.016$ \\
%& AG-CUResNeSt-101 \cite{agcuresnest} & $0.946 \pm 0.010$ & $0.902 \pm 0.015$ & $0.953 \pm 0.013$ & \textbf{0.944 $\pm$ 0.009} \\
%\cline{2-6}
& \textit{\textbf{\ModelName-S~(Ours)}} & \textbf{0.948 $\pm$ 0.002} & \textbf{0.904 $\pm$ 0.004} & \textbf{0.958 $\pm$ 0.003} & \textbf{0.941 $\pm$ 0.004} \\
& \textit{\textbf{\ModelName-L~(Ours)}} & \underline{\textit{0.947 $\pm$ 0.002}} & \underline{\textit{0.903 $\pm$ 0.003}} & \underline{\textit{0.956 $\pm$ 0.002}} & \underline{\textit{0.942 $\pm$ 0.005}}\\[2pt]
\hline
\hline
\multirow{8}{*}{\rotatebox[origin=c]{90}{Kvasir}} 
& UNet \cite{unet}                 & $0.708\pm0.017$          & $0.602\pm0.010$         & $0.805\pm0.014$  & $0.716\pm0.020$      \\
& ResUNet++ \cite{resunet++}            & $0.780\pm0.010$          & $0.681\pm0.008$         & $0.834\pm0.010$  & $0.799\pm0.010$      \\
& DoubleUNet \cite{doubleunet} & $0.879 \pm 0.018$ & $0.816 \pm 0.026$ & $0.902 \pm 0.027$ & $0.894 \pm 0.039$ \\
& DDANet \cite{ddanet} & $0.860 \pm 0.005$ & $0.791 \pm 0.004$ & $0.876 \pm 0.015$ & $0.892 \pm 0.018$ \\
& ColonSegNet \cite{colonsegnet} & $0.676 \pm 0.037$ & $0.557 \pm 0.040$ & $0.731 \pm 0.088$ & $0.730 \pm 0.080$ \\
& HarDNet-MSEG \cite{hardnet_mseg} & $0.889 \pm 0.011$ & $0.831 \pm 0.011$ & $0.892 \pm 0.015$ & $0.926 \pm 0.014$ \\
& PraNet \cite{pranet} & $0.883\pm0.020$          & $0.822\pm0.020$         & $0.897\pm0.020$  & $0.906\pm0.010$      \\
%& AG-CUResNeSt-101 \cite{agcuresnest}) & $0.912\pm0.010$ & $0.860\pm0.011$       & $0.923\pm0.009$  & \textbf{0.927 $\pm$ 0.014}      \\
%\cline{2-6}
& \textit{\textbf{\ModelName-S~(Ours)}} & \textbf{0.924 $\pm$ 0.008} & \textbf{0.875 $\pm$ 0.010} & \textbf{0.941 $\pm$ 0.010} & \textbf{0.927 $\pm$ 0.008} \\
& \textit{\textbf{\ModelName-L~(Ours)}} & \underline{\textit{0.917 $\pm$ 0.006}} & \underline{\textit{0.865 $\pm$ 0.007}} & \underline{\textit{0.932 $\pm$ 0.007}} & \underline{\textit{0.926 $\pm$ 0.008}} \\[2pt]
\hline
\end{tabular}
}
\label{tab:kfold}
\end{table*}

\begin{figure*}[ht!]
\centering
\begin{subfigure}[b]{0.49\textwidth}
\centering
\includegraphics[width=\textwidth]{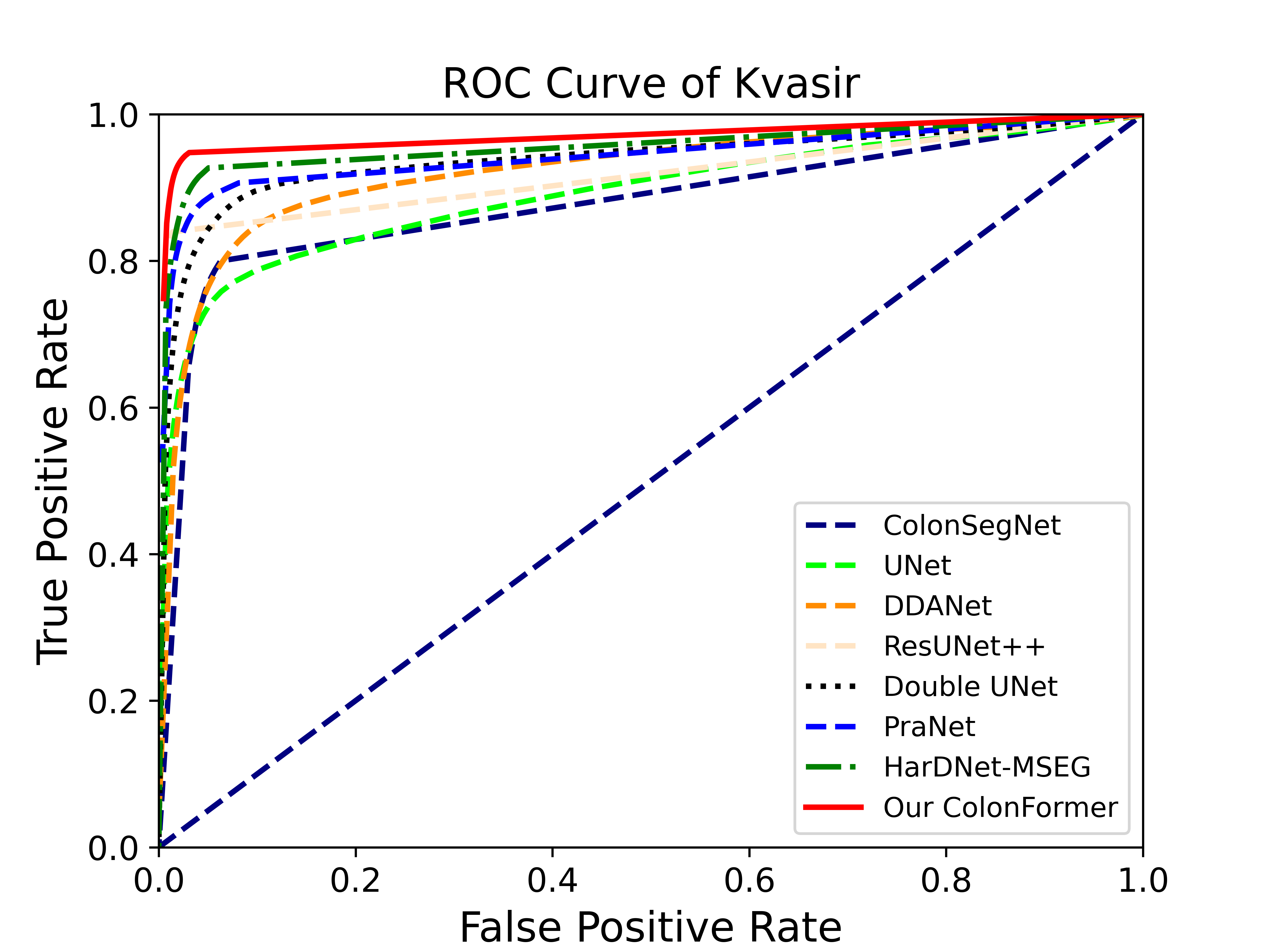}
\caption{ROC curves}
\end{subfigure}
\hfill
\begin{subfigure}[b]{0.49\textwidth}
\centering
\includegraphics[width=\textwidth]{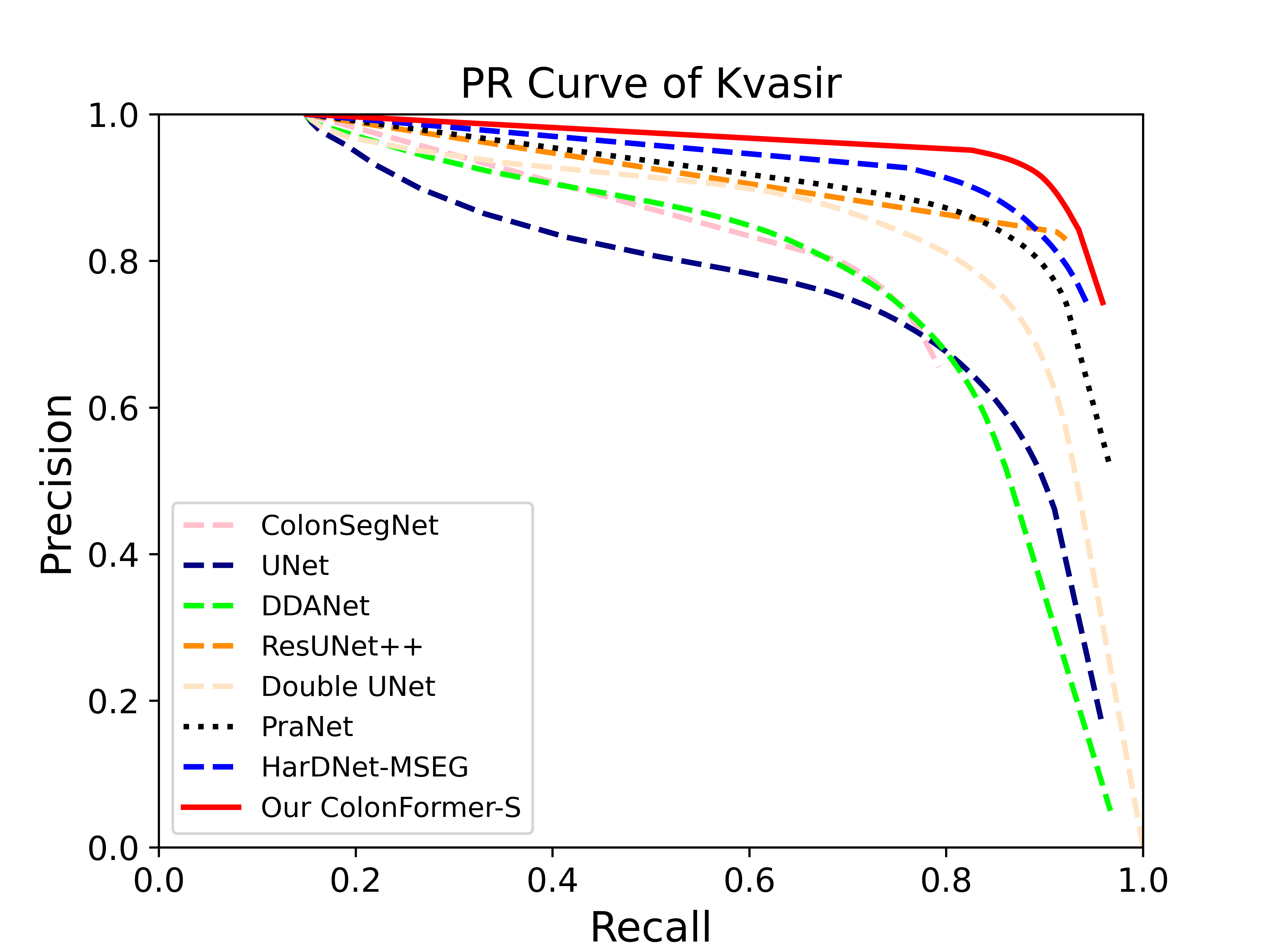}
\caption{PR curves}
\end{subfigure}
\caption{ROC curves and PR curves  on the 5-fold cross-validation on the Kvasir-SEG dataset. All the curves are averaged over 5 folds.}
\label{fig:kvasir_ROC}
\end{figure*}

\begin{figure*}[ht!]
\centering
\includegraphics[width=\textwidth]{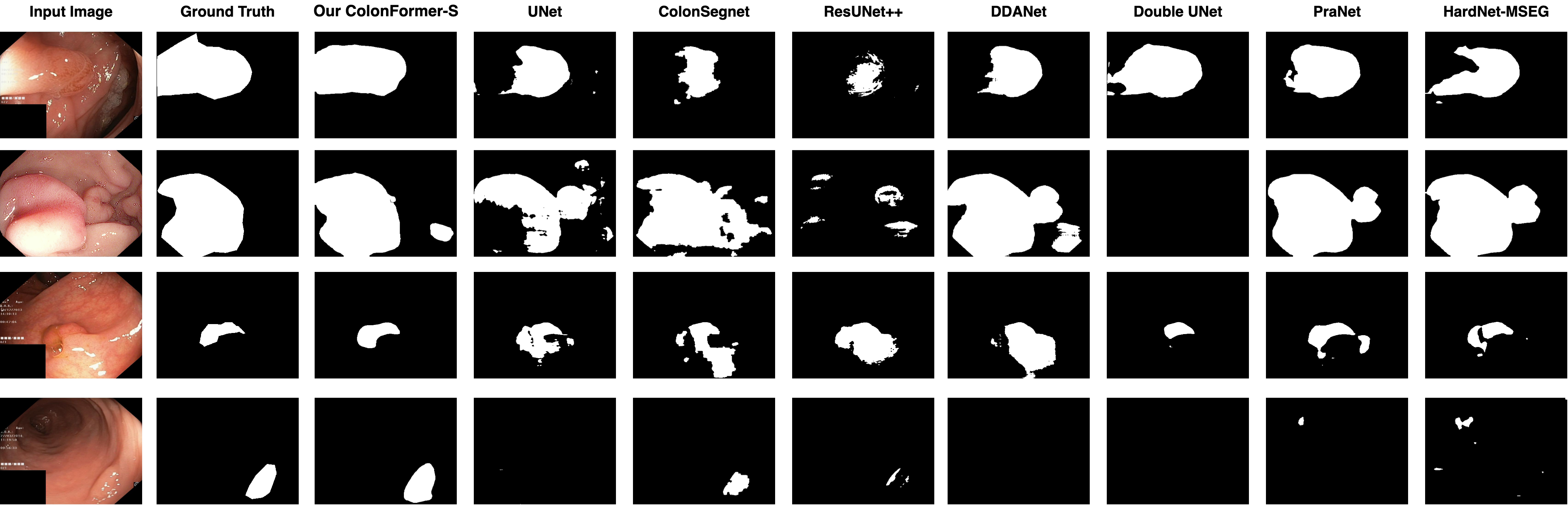}
\caption{Qualitative result comparison of different models trained on the first fold of the 5-fold cross-validation on the Kvasir dataset.}
\label{fig:kvasir_kfold}
\end{figure*}

\subsection{Datasets}
We perform experiments on the five popular datasets for polyp segmentation: Kvasir \cite{kvasir}, CVC-Clinic DB \cite{cvc_clinic}, CVC-Colon DB \cite{cvc_colon}, CVC-T \cite{endo}, and ETIS-Larib Polyp DB \cite{ETIS}. Details of these datasets are described as follows:

\begin{itemize}
\item \textbf{Kvasir dataset} is collected using endoscopic equipment at Vestre Viken Health Trust (VV), Norway. Images are carefully annotated and verified by experienced gastroenterologists from VV and the Cancer Registry of Norway. The dataset consists of 1000 images with different resolutions from $720\times576$ to $1920\times1072$ pixels.
\item \textbf{CVC-ClinicDB dataset} is a database of frames extracted from colonoscopy videos. The dataset consists of 612 images with a resolution of $384\times288$ pixels from 31 colonoscopy sequences. The dataset was used in the training stages of the MICCAI 2015 Sub-Challenge on Automatic Polyp Detection Challenge in Colonoscopy Videos.
\item \textbf{CVC-ColonDB dataset} is provided by the Machine Vision Group (MVG). The dataset consists of 380 images with a resolution of $574\times500$ pixels from 15 short colonoscopy videos.
\item \textbf{CVC-T dataset} is the test set of a more extensive dataset called Endoscene. CVC-T consists of 60 images obtained from 44 video sequences acquired from 36 patients.

\item \textbf{ETIS-Larib dataset} contains 196 high resolution ($1226\times996$) colonoscopy images.
\end{itemize}

\begin{table*}[ht!]
\caption{Performance comparison of different methods on cross-dataset configurations. All results are averaged over five runs.}
\centering
{\renewcommand{\arraystretch}{1.2}
\begin{tabular}{cc|c|c|cccc}
\hline
\multicolumn{2}{c|}{Train} & Test & Method & mDice & mIOU  & Recall & Precision  \\
\hline
\hline
\multirow{7}{*}{\rotatebox[origin=c]{90}{CVC-ColonDB}} & \multirow{7}{*}{\rotatebox[origin=c]{90}{+ ETIS-Larib}}  &  \multirow{7}{*}{\rotatebox[origin=c]{90}{CVC-ClinicDB}} 
& ResUNet++ \cite{resunet++} & 0.406 & 0.302 & 0.481 & 0.496 \\
&& & ColonSegNet \cite{colonsegnet} & 0.427 & 0.321 & 0.529 & 0.552 \\
&& & DDANet \cite{ddanet} & 0.624 & 0.515 & 0.697 & 0.692 \\
&& & DoubleUNet \cite{doubleunet} & 0.738 &  0.651 & 0.758 & 0.824 \\
&& & HarDNet-MSEG \cite{hardnet_mseg} & 0.765 &  0.681 & 0.774 & 0.863 \\
&& & PraNet \cite{pranet} & 0.779 &  0.689 & 0.832 & 0.812 \\
%& & AG-CUResNeSt-101 \cite{agcuresnest} & 0.833 & 0.754 & 0.840 & 0.883  \\
%\cline{3-7}
&& & \textit{\textbf{\ModelName-S~(Ours)}} & \textbf{0.851}  &   \textbf{0.771}  &   \textbf{0.853}   &  \underline{\textit{0.896}} \\
&& & \textit{\textbf{\ModelName-L~(Ours)}}     &   \underline{\textit{0.847}}  &   \underline{\textit{0.770}}  &   \underline{\textit{0.844}}   &  \textbf{0.902} \\[2pt]
\hline
\hline
\multirow{9}{*}{\rotatebox[origin=c]{90}{CVC-ColonDB}} & &  \multirow{9}{*}{\rotatebox[origin=c]{90}{CVC-ClinicDB}} 
& ResUNet++ \cite{resunet++} & 0.339 & 0.247 & 0.380 & 0.484 \\
&& & DoubleUNet \cite{doubleunet} & 0.441 &  0.375 & 0.423 & 0.639 \\
&& & DDANet \cite{ddanet} & 0.476 & 0.370 & 0.501 & 0.644 \\
%& ResNet50-Mask-RCNN \cite{maskrcnn} & 0.639    & 0.560        & 0.648  & 0.710      \\
&& & ResNet101-Mask-RCNN \cite{maskrcnn} & 0.641 & 0.565        & 0.646  & 0.725      \\
&& & ColonSegNet \cite{colonsegnet} & 0.582 &    0.268 & 0.511 & 0.460 \\
&& & HarDNet-MSEG \cite{hardnet_mseg} & 0.721 &  0.633 & 0.744 & 0.818 \\
&& & PraNet \cite{pranet} & 0.738 & 0.647 & 0.751 & 0.832 \\
%& & AG-CUResNeSt-101 \cite{agcuresnest}   & 0.771  & 0.686        & 0.793  & 0.830  \\
%\cline{3-7}
&& & \textit{\textbf{\ModelName-S~(Ours)}} & \textbf{0.816}    &   \textbf{0.731} & \textbf{0.809}  & \textbf{0.881} \\
&& & \textit{\textbf{\ModelName-L~(Ours)}} & \underline{\textit{0.804}}    &   \underline{\textit{0.723}} & \underline{\textit{0.794}}  & \underline{\textit{0.877}}\\[2pt]
\hline
\hline
\multirow{10}{*}{\rotatebox[origin=c]{90}{CVC-ClinicDB}} & &  \multirow{10}{*}{\rotatebox[origin=c]{90}{ETIS-Larib}} 
& ResUNet++ \cite{resunet++} & 0.211 & 0.155 & 0.309 & 0.203 \\
&& & ColonSegNet \cite{colonsegnet} & 0.217 &    0.110 & 0.654 & 0.144 \\
%&&&  ResNet50-Mask-RCNN \cite{maskrcnn} & 0.501   & 0.412         & 0.546  & 0.573      \\
&& & DDANet \cite{ddanet} & 0.400 & 0.313 & 0.507 & 0.464 \\
&& & ResNet101-Mask-RCNN \cite{maskrcnn} & 0.565   & 0.469         & 0.565  & 0.639      \\
&& & DoubleUNet \cite{doubleunet} & 0.588   & 0.500         & 0.689  & 0.599      \\
&& & PraNet \cite{pranet} & 0.631 &    0.555 &  0.762  &    0.597\\
&& & HarDNet-MSEG \cite{hardnet_mseg} & 0.659 &  0.583 & 0.676 & 0.705 \\
%& & AG-CUResNeSt-101 \cite{agcuresnest} & 0.701  & 0.613         & 0.755  & 0.693      \\
%\cline{3-7}
&& & \textit{\textbf{\ModelName-S~(Ours)}} & \underline{\textit{0.723}}    &   \underline{\textit{0.635}} & \underline{\textit{0.797}}  & \underline{\textit{0.731}}\\
&& & \textit{\textbf{\ModelName-L~(Ours)}}  & \textbf{0.760}    &   \textbf{0.673} & \textbf{0.859}  & \textbf{0.734}\\[2pt]
\hline
\end{tabular}
}
\label{tab:cross-dataset}
\end{table*}

\subsection{Experiment settings}
We implement \ModelName using the PyTorch framework. For a fair setting and comparison, we use the same parameters as \cite{segformer} for the MiT backbone: kernel size $K=7$, stride $S=4$, padding size $P=3$, and $K=3, S=2, P=1$ to produce features with the same size as the non-overlapping process. Based on experiments in \cite{f3net}, \cite{focalloss}, we use $\lambda=5, \alpha=0.25$ and $\gamma=2$ for the losses in Eq.~(\ref{eq:wfocal}) and Eq.~(\ref{eq:wiou}). Training is performed using Google Colab on virtual machines with 16GB RAM and an NVIDIA Tesla P100 GPU. Input images are resized to $352\times352$ for testing. In order to increase the model's robustness w.r.t image sizes, the training images are consequently scaled with a factor of $\{0.75, 1, 1.25\}$, respectively,  and fed to the model for learning. None of the data augmentation techniques is used in the training phase. 

We use six experiment setups to evaluate our method; each setup is described in detail below: 
\begin{itemize}
\item \textbf{Experiment 1}: We use the same split as suggested in \cite{pranet}, where $90\%$ of the Kvasir and ClinicDB datasets are used for training. The remaining images in the Kvasir and CVC-ClinicDB datasets and all images from CVC-ColonDB, CVC-T, and ETIS-Larib are used for testing. 
\item \textbf{Experiment 2}:~5-fold cross-validation on the CVC-ClinicDB and Kvasir datasets. Each dataset is divided into five equal folds. Each run uses one fold for testing and four remaining folds for training.
\item \textbf{Experiment 3}: Cross-dataset evaluation with 3 training-testing configurations:
\begin{enumerate}
\item CVC-ColonDB and ETIS-Larib for training, CVC-ClinicDB for testing;
\item CVC-ColonDB for training, CVC-ClinicDB for testing;
\item CVC-ClinicDB for training, ETIS-Larib for testing.
\end{enumerate}
\end{itemize}

The first experiment compares our \ModelName model with state-of-the-art CNN-based and Transformer-based networks using the same widely-used dataset configuration as suggested in \cite{pranet}.
The second experiment compares \ModelName's learning ability to several recent polyp segmentation methods. Finally, the last experiment provides deeper insights into the generalization capability of \ModelName and other benchmark models. 

We use the Adam optimizer and cosine annealing scheduler with a learning rate of 1e-4. \ModelName is trained in 20 epochs with a batch size of 8. The checkpoint of the last epoch is used for evaluation. Except for the second experiment with 5-fold cross-validation, we train \ModelName five times, and the \ModelName's results are averaged over five runs.

In addition, we perform a series of ablation studies to evaluate the effectiveness of each component in the proposed \ModelName. All ablation studies are performed on the dataset configuration for Experiment 1.
% \begin{itemize}
% \item Investigating the effectiveness of the UPer Decoder
% \item Investigating the effectiveness of the Refinement Module
% \item Investigating the effectiveness of the Mix Transformer backbone
% \end{itemize}

\subsection{Comparison with benchmark models}
Table \ref{tab:sota} describes the comparison results for Experiment 1. \ModelName generally outperforms the benchmark models on most datasets. Notably, both \ModelName-S and \ModelName-L outperform the second-best TransFuse-L* by $3\%$ in mDice and $2.7\%$ in mIOU on the ColonDB dataset. Compared to the second-best CaraNet on the ETIS-Larib dataset, \ModelName-S achieves an improvement of $5.2\%$ in mDice, and $4.8\%$ in mIOU, while \ModelName-L achieves an improvement of $6.4\%$ in mDice and $5.9\%$ in mIOU. The high capacity of \ModelName-L seems more suitable for the high resolution of images in the ETIS-Larib dataset. 
However, both \ModelName-S and \ModelName-L achieve roughly $1\%$ lower metrics against TransFuse-L* on the CVC-ClinicDB dataset, whose images obtain very low resolution.

Table \ref{tab:kfold} describes the comparison results for Experiment 2. We report both the average value and the standard deviation for each metric, which reflects the models' stability. One can see that both \ModelName-S and \ModelName-L outperform all other state-of-the-art models in mDice, mIOU, precision, and recall on both datasets. Notably, our \ModelName is the most stable model on both datasets, achieving the lowest standard deviation for each evaluation metric.

Qualitative results for Experiment 2 are shown in Fig.~\ref{fig:kvasir_kfold}. \ModelName-S demonstrates much fewer wrongly predicted pixels in segmentation results than other models. \ModelName-S also produces better ROC and PR curves than the benchmark models, as depicted in  Fig.~\ref{fig:kvasir_ROC}.

Table \ref{tab:cross-dataset} describes the comparison results for Experiment 3. Overall, both \ModelName-S and \ModelName-L significantly outperform benchmark models on cross-dataset metrics. For the first configuration, \ModelName-S outperforms the second-best PraNet by $7-8\%$ on all metrics. In the second configuration, \ModelName-S continues to achieve a $5.8\%$ improvement in precision and $7.8\%$ improvement in mDice over PraNet. For the third configuration, \ModelName-L again shows its suitability to the ETIS-Larib dataset achieving a $10.1\%$ improvement in mDice and $18.3\%$ in recall over PraNet. These are highly significant improvements, showing that our \ModelName can generalize very well to new unseen data. Some result samples for this experiment are shown in Fig.~\ref{fig:colon_clinic}. Similar to Fig.~\ref{fig:kvasir_kfold}, one can see that \ModelName yields better segmentation results compared to other state-of-the-arts.

\begin{figure*}[ht!]
\centering
\includegraphics[width=\textwidth]{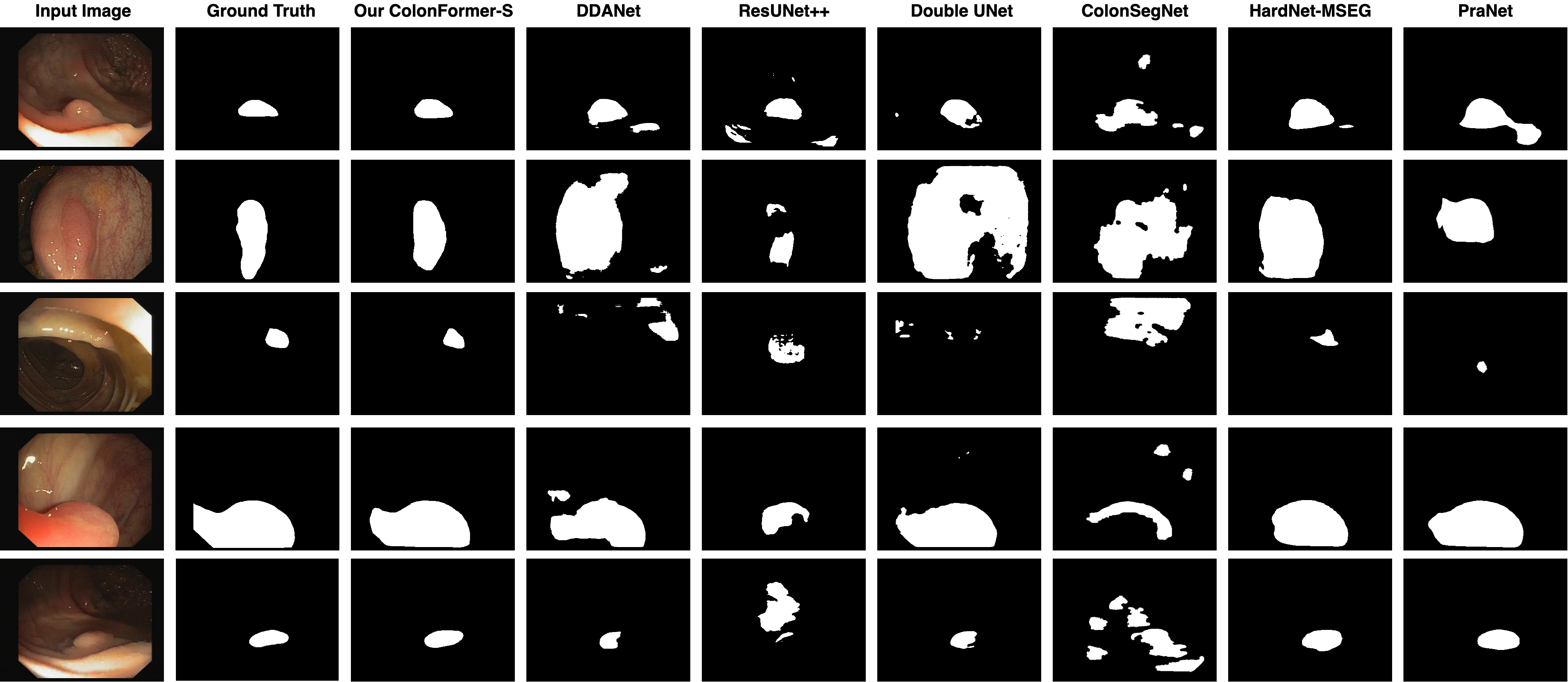}
\caption{Qualitative result comparison using CVC-Colon for training and CVC-Clinic for testing.}
\label{fig:colon_clinic}
\end{figure*}

Table~\ref{tab:complexity} compares \ModelName with other benchmark models in terms of size and computational complexity. One can see that our \ModelName-S obtains competitive size and computational complexity compared to the most lightweight CNN-based models such as PraNet \cite{pranet}, and HarDNet-MSEG \cite{hardnet_mseg}. Our \ModelName-L is larger than most CNN-based neural networks but still more efficient than other Transformer-based methods in terms of GFlops.

\begin{table*}[ht!]
\caption{Number of parameters and GFLOPs of different methods}
\centering
{\renewcommand{\arraystretch}{1.2}
\begin{tabular}{c|cc}
\hline
Method & Parameters (M) & GFLOPs\\
\hline
\hline
PraNet \cite{pranet}  & 32.55 &   13.11        \\
HarDNet-MSEG \cite{hardnet_mseg} & 33.34 & 11.38         \\
CaraNet \cite{caranet} & 46.64 & 21.69       \\
TransUNet \cite{transunet} & 105.5 & 60.75         \\
%TransFuse-S & - & -         \\
%TransFuse-L & - & -         \\
TransFuse-L* \cite{transfuse} & - & -          \\
SegFormer-B3 \cite{segformer} & 47.22 & 33.68 \\
SegFormer-B3-Uper & 46.61 & 20.99 \\
\textit{\textbf{\ModelName-S~(Ours)}} & 33.04 & 16.03 \\
%\hline
%\ModelName 3blocks & 52.59 & 18.02 \\
%\ModelName 4blocks & 53.04 & 23.67 \\
\textit{\textbf{\ModelName-L~(Ours)}} & 52.94 & 22.94         \\
\hline
\end{tabular}
}
\label{tab:complexity}
\end{table*}
\subsection{Ablation studies}
\label{sec:ablation_studies}
\textbf{Effectiveness of the UPer Decoder.} We firstly compare the original SegFormer-B3 \cite{segformer} with MLP Decoder and another model called SegFormer-B3-Uper that replaces the original MLP decoder with the UPer Decoder. Both models use the MiT-B3 backbone in terms of encoder. 
%\textcolor{red}{(Oanh: can chi ro cac khoi khac nhu the nao va su dung setting nao. Thông tin tren caption cac bang cung nen cu the dieu do - Comment nay tuong tu cho 2 phan sau ve Refinement Module va Backbone)}. 
Results are shown in the first two rows of Table~\ref{tab:ablation}. Both network versions show similar metrics across the test datasets, with slight variations of roughly $1\%$. However, one can see from Table~\ref{tab:complexity} that UPer Decoder is also significantly less costly, requiring only 20.99 GFLOPs as opposed to MLP Decoder (33.68 GFLOPs). These results compel us to choose the UPer decoder for \ModelName, which alleviates the large computation costs incurred with the Transformer backbone.
%\textcolor{green}{Oanh: 2 bang nho o Table~\ref{tab:decoder} nen ghep lai, de thanh 2 cot lon: 1 ve performance measure, 2 ve computional cost\\}

\textbf{Effectiveness of the Refinement Module.} We evaluate the performance of SegFormer-B3-Uper-ARA with the {{A-RA}} Refinement Module as in \cite{caranet}, and our \ModelName-L with the adjusted Refinement Module as described in Section \ref{sec:refinement_module}. Results are shown in Table~\ref{tab:ablation}. Overall, incorporating the Refinement Module yields improvement across all datasets. Our \ModelName-L also yields superior performance than SegFormer-B3-Uper-ARA on the Kvasir, CVC-ClinicDB, and most significantly, the ETIS-Larib datasets, while slightly underperforming on the CVC-ColonDB and CVC-T datasets.

\textbf{Effectiveness of the MiT Backbone.} The Mix Transformer (MiT) \cite{segformer} backbone has several variations ranging from MiT-B0 to MiT-B5. Accordingly, our \ModelName also have different variations, including \ModelName-XS, \ModelName-S, \ModelName-L, \ModelName-XL, \ModelName-XXL, respectively. Table~\ref{tab:backbone} shows our comparison between all variations of \ModelName. Overall, \ModelName-S and \ModelName-L yield the best average results across our test datasets.

\begin{table*}[ht!]
\caption{Ablation study on the effectiveness of different components. All results are averaged over five runs.}
\centering
{\renewcommand{\arraystretch}{1.2}
\resizebox{\textwidth}{!}{
\begin{tabular}{c|ccc|cc|cc|cc|cc|cc}
\hline
Method & Uper & A-RA & RA-RA & \multicolumn{2}{c|}{Kvasir} & \multicolumn{2}{c|}{CVC-ClinicDB} & \multicolumn{2}{c|}{CVC-ColonDB} & \multicolumn{2}{c|}{CVC-T} & \multicolumn{2}{c}{ETIS-Larib}  \\
\cline{5-14}
&&&& mDice & mIOU               & mDice & mIOU                 & mDice & mIOU                & mDice & mIOU                   & mDice & mIOU              \\
\hline
\hline
SegFormer-B3 \cite{segformer} & \textemdash &  \textemdash & \textemdash           & 0.920 & 0.866 & \underline{\textit{0.925}} & \underline{\textit{0.876}} & 0.806 & 0.726 & 0.905 & 0.840  & \underline{\textit{0.786}} & \underline{\textit{0.707}} \\
%SegFormer-Uper  & \checkmark &  \textemdash & \textemdash           & 0.921 & 0.869 & 0.928 & 0.881 & 0.795 & 0.718 & 0.904 & 0.838  & 0.782 & 0.704 \\
SegFormer-B3-Uper & \checkmark &  \textemdash & \textemdash           & 0.916 & 0.864 & 0.924 & 0.876 & \underline{\textit{0.811}} & 0.731 & 0.900 & 0.832  & 0.784 & \underline{\textit{0.707}} \\
% SegFormer-Uper-ARA & \checkmark   &  \checkmark & \textemdash         & 0.921 & 0.871 & 0.928 & 0.881 & \textbf{0.815} & \textbf{0.737} & \textbf{0.912} & \textbf{0.848}  & 0.787 & 0.708 \\
SegFormer-B3-Uper-ARA & \checkmark   &  \checkmark & \textemdash         & \underline{\textit{0.922}} & \underline{\textit{0.872}} & 0.922 & 0.875 & \textbf{0.812} & \textbf{0.734} & \underline{\textit{0.903}} & \underline{\textit{0.837}}  & 0.787 & 0.704 \\

\textit{\textbf{\ModelName-L~(Ours)}} & \checkmark   &  \textemdash  & \checkmark  & \textbf{0.924} & \textbf{0.876} & \textbf{0.932} & \textbf{0.884} & \underline{\textit{0.811}} & \underline{\textit{0.733}} & \textbf{0.906} & \textbf{0.842} & \textbf{0.801} & \textbf{0.722} \\[2pt]
\hline
\end{tabular}
}
}
\label{tab:ablation}
\end{table*}

\begin{table*}[ht!]
\caption{Evaluation metrics for different variations of the MiT backbone. All results are averaged over five runs.}
\centering
{\renewcommand{\arraystretch}{1.2}
\resizebox{\textwidth}{!}{
\begin{tabular}{c|c|cc|cc|cc|cc|cc}
\hline
Method & Backbone & \multicolumn{2}{c|}{Kvasir} & \multicolumn{2}{c|}{CVC-ClinicDB} & \multicolumn{2}{c|}{CVC-ColonDB} & \multicolumn{2}{c|}{CVC-T} & \multicolumn{2}{c} {ETIS-Larib} \\ % & \multicolumn{2}{c} {All five datasets}  \\
\cline{3-12} 
&& mDice & mIOU               & mDice & mIOU                 & mDice & mIOU                & mDice & mIOU                   & mDice & mIOU  \\ %& mDice & mIOU            \\
\hline
\hline
\ModelName-XS & MiT-B1           & 0.913 & 0.859 & 0.926 & 0.876 & 0.784 & 0.700 & 0.879 & 0.808  & 0.758 & 0.679 \\ %& 0.812 &	0.737 \\
\ModelName-S & MiT-B2           & \textbf{0.927} & \textbf{0.877} & \textbf{0.932} & \underline{\textit{0.883}} & \underline{\textit{0.811}} & 0.730 & 0.894 & 0.826  & 0.789 & 0.711 \\ %& 0.835 &	0.763 \\
\ModelName-L & MiT-B3           & \underline{\textit{0.924}} & \underline{\textit{0.876}} & \textbf{0.932} & \textbf{0.884} & \underline{\textit{0.811}} & \underline{\textit{0.733}} & \textbf{0.906} & \textbf{0.842} & \textbf{0.801} & \textbf{0.722} \\ %& \textbf{0.839} &	\textbf{0.768}\\
\ModelName-XL & MiT-B4           & 0.920 & 0.870 & 0.923 & 0.875 & \textbf{0.814} & \textbf{0.735} & \underline{\textit{0.905}} & \underline{\textit{0.840}}  & \underline{\textit{0.795}} & \underline{\textit{0.715}} \\ %&  \underline{\textit{0.838}}	& \underline{\textit{0.765}}\\
\ModelName-XXL & MiT-B5           & 0.920 & 0.872 & 0.924 & 0.876 & 0.802 & 0.724 & 0.899 & 0.831 & 0.776 & 0.700 \\ % & 0.827 &	0.756\\
\hline
\end{tabular}
}
}
\label{tab:backbone}
\end{table*}

%\textcolor{red}{Oanh: nen chon loc bo sung ket qua truc quan de minh hoa cho phuong phap khi so sanh voi pp khac, hoac/va khi nhan vao tam quan trong cua 1 so module co dong gop noi troi}

\section{Conclusion}
\label{sec:conclude}
This paper proposes a novel deep neural network architecture called \ModelName for colon polyp segmentation. Our model leverages both the advantages of Transformers and CNNs architectures to learn a powerful multi-scale hierarchical feature representation. We also enhance the reverse attention with axial attention by relaxing it with a residual connection. The refinement module allows the network to incrementally correct the polyp boundary from a coarse global map produced by the decoder. Our extensive experiments show that \ModelName significantly outperforms existing state-of-the-art models on popular benchmark datasets.

In future works, we will investigate lightweight or sparse self-attention layers to reduce the computational complexity. In addition, other types of architectures for combining Transformers and CNNs can also be exploited.

%\textcolor{red}{Oanh:\\
%- bo sung future works: co the dua ra 1 vai diem yeu cua ket qua hien tai va dinh huong giai quyet trong tuong lai}

\section{Acknowledgments}
This work was funded by Vingroup Innovation Foundation
(VINIF) under project code VINIF.2020.DA17.


%   \bibitem{kour2014real}
%   George Kour and Raid Saabne.
%   \newblock Real-time segmentation of on-line handwritten arabic script.
%   \newblock In {\em Frontiers in Handwriting Recognition (ICFHR), 2014 14th
%           International Conference on}, pages 417--422. IEEE, 2014.

%   \bibitem{kour2014fast}
%   George Kour and Raid Saabne.
%   \newblock Fast classification of handwritten on-line arabic characters.
%   \newblock In {\em Soft Computing and Pattern Recognition (SoCPaR), 2014 6th
%           International Conference of}, pages 312--318. IEEE, 2014.

%   \bibitem{hadash2018estimate}
%   Guy Hadash, Einat Kermany, Boaz Carmeli, Ofer Lavi, George Kour, and Alon
%   \newblock Estimate and replace: A novel approach to integrating deep neural
%   networks with existing applications.
%   \newblock {\em arXiv preprint arXiv:1804.09028}, 2018.

% \end{thebibliography}

\bibliographystyle{unsrt}
\bibliography{arxiv}

\end{document}